# Lexical and Statistical Analysis of Bangla Newspaper and Literature: A Corpus-Driven Study on Diversity, Readability, and NLP Adaptation

Pramit Bhattacharyya[1]    Arnab Bhattacharya[1]

[1] IIT Kanpur
{pramitb, arnabb}@cse.iitk.ac.in

# Contents

**Abstract**

In this paper, we present a comprehensive corpus-driven analysis of Bangla literary and newspaper texts to investigate their lexical diversity, structural complexity and readability. We undertook Vācaspati and IndicCorp, which are the most extensive literature and newspaper-only corpora for Bangla. We examine key linguistic properties, including the type-token ratio (TTR), hapax legomena ratio (HLR), Bigram diversity, average syllable and word lengths, and adherence to Zipf's Law, for both newspaper (IndicCorp) and literary corpora (Vācaspati). Our findings reveal that till $\sim 10^6$ words, the newspaper corpora show a higher TTR than the literary corpora; after that, the literary corpora have a higher TTR ratio. For all other features, such as Bigram Diversity and HLR, despite its smaller size, the literary corpus exhibits significantly higher lexical richness and structural variation. We have also demonstrated that a literary corpus adheres more closely to global word distribution properties, such as Zipf's law, than a newspaper corpus or a merged corpus of both literary and newspaper texts. Literature corpora also have higher entropy and lower redundancy values compared to a newspaper corpus. We also further assess the readability using Flesch and Coleman-Liau indices, showing that literary texts are more complex. Additionally, we tried to understand the diversity of corpora by building $n$-gram models and measuring perplexity. Our findings reveal that literary corpora have higher perplexity than newspaper corpora, even for similar sentence sizes. This trend can also be observed for the English newspaper and literature corpus, indicating its generalizability. We also examined how the performance of models on downstream tasks is influenced by the inclusion of literary data alongside newspaper data. Our findings suggest that integrating literary data with newspapers improves the performance of models on various downstream tasks.

# 1 Introduction

Bangla is the sixth-most spoken language in the world, yet most large-scale corpora for Bangla are curated from web text [22] or newspapers [8, 13, 15]. While these sources are relatively easier to collect, they exhibit stylistic simplicity and lack the lexical diversity required to capture the morphological richness of the language. In contrast, literary texts encompass a broader range of vocabulary, syntax, and discourse structure. IndicCorp v2 [6] and BanglaBert [2] contain Bangla literary data in addition to newspaper data and other web data. In our previous work, we curated Vācaspati, a literary-only dataset on Bangla.

In our previous work [3], we demonstrate that models developed on top of intelligently curated Vācaspati corpora of only literary data can yield state-of-the-art results even with limited resources. In this chapter, we try to delve deep into the lexical richness of the Vācaspati, a literary-only corpus, compared to IndicCorp, a newspaper-only corpus, as well as their combination in a merged setting. Similar work has been done for Hindi and Sanskrit [12], however our work is more detailed.

We begin by quantifying basic lexical statistics, such as average word and sentence lengths, unique word counts, type–token ratios, hapax legomena ratios, and word bigram diversity, across the corpora. These analyses enable us to investigate whether a smaller, yet lexically richer, literary corpus can yield more varied patterns than a much larger newspaper corpus.

We further studied word distribution properties of the corpora through Zipf's law [23], entropy and redundancy [21] to validate that an intelligently curated corpora is more information rich compared to a large newspaper corpora. Additionally, we built $n$-gram models (uni, bi, tri) on top of Vācaspati and IndicCorp and measured the perplexity [11] of the models. The results indicate that models built on top of Vācaspati are less confident in their predictions compared to models built on top of IndicCorp. We further conducted the same experiment with an English newspaper and a literary dataset and observed that this phenomenon generalizes across languages.

Finally, we go beyond intrinsic corpus statistics and evaluate downstream impact, first by using lexicons derived from each corpus for real-word merge detection task, and then by pretraining Electra-style models on different mixtures of Vācaspati and IndicCorp and testing them on multiple Bangla tasks.

In summary, our contributions are briefly described below:-

- We studied surface-level lexical features, such as type-to-token ratio, hapax-legomena ratio, and word bigram diversity, to demonstrate the lexical diversity of Vācaspati compared to IndicCorp dataset (Section 3).

- We further studied the *readability* of each corpora to understand which corpora is more difficult to read using established techniques like **Flesch Reading Ease** (Section 7).

- We then studied the global word distribution properties of each corpus through Zipf's law, entropy and redundancy properties (Sec 5). Additionally, we tried to understand the diversity of corpora by building $n$-gram models and measuring perplexity. We also assessed whether this property is generalized or is specific to Bangla (Sec 6).

- We finally evaluated the performance of corpora on downstream tasks for real-word merge detection and then by pretraining Electra-style models on different mixtures of Vācaspati and IndicCorp and testing them on multiple Bangla tasks (Section 8).

## 2 Related Work

**Corpora:** Most Bangla datasets to date have been generated from newspaper articles. The EMILLE-CIIL [15] monolingual written corpus consists of 1,980,000 words from newspapers, whereas the spoken corpus consists of only 442,000 words. Wikipedia dump of Bangla is of size 326 MB (as on July 1, 2023). The Leipzig corpus [8] on Bangla, curated by crawling newspapers, consists of 1,200,255 sentences and 16,632,554 tokens. The SUPara corpus [16] is an English-Bangla sentence-aligned parallel corpus consisting of more than 200,000 words. OPUS [22], a multilingual corpus of translated open-source documents, curated a sentence-aligned parallel corpus for Bangla with 4.7 million tokens and 0.51 million sentences. The BNLP [19] dataset for embedding was curated by collecting 127,867 news articles and Wikipedia dumps, and was trained on 5.83 million sentences and 93.43 million tokens. Hasan et.al [9] built a sentence-aligned Bangla-English parallel corpus with 2.75M sentences focusing mainly on machine translation work. The IndicCorp corpus [13] for Bangla consists of 39.9 million sentences and 836 million tokens generated from 3.89 million news articles. IndicCorp v2 [6] generated from newspaper articles, e-books and other webpages contains 936 million tokens in Bangla. The OSCAR project [17] built by crawling websites using CommonCrawl (https://commoncrawl.org) contains 632 million words for Bangla. VĀCASPATI [3] is a literature-only data for Bangla. In this work we compared lexical diversity of only VĀCASPATI and IndicCorp (largest newspaper dataset in Bangla). Similar work has been done for Hindi and Sanskrit [12], however our work is more detailed.

## 3 Lexical Analysis

Lexical analysis is essential for Natural Language Processing (NLP) because it helps understand vocabulary diversity, sentence structure, and syntactic complexity. Comparing newspaper and literary corpora is particularly significant, as they differ considerably in terms of word richness, structure, and lexical variety. In this section, we compare the VĀCASPATI (Literary) Corpus [3] and the IndicCorp (Newspaper) corpus [13].

A larger corpus generally provides a better statistical representation of language patterns, which is the driving factor behind modern-day LLMs. The **IndicCorp** is way larger (*844M+ words and 63.8M+ sentences*) than **Vācaspati** (*102M+ words and 11.95M+ sentences*). From the outset, a newspaper corpus is always better than a literary corpus because of its larger size and ease of availability. In this section, we will try to validate that even a smaller literary corpus is way more lexically enriched and diverse than a large newspaper corpus. The results of the metrics used for this analysis are demonstrated in Table 1.

### 3.1 Average Word Length

Average word length refers to the *average number of characters per word* in a corpus. It is an important *morphological feature* in computational linguistics, as it helps in analyzing *word complexity, readability, and structural variations* of a corpus. Average character length is represented by:

$$\text{Avg. Word Length} = \frac{\text{Total Characters in all words}}{\text{Total Words}} \quad (1)$$

VĀCASPATI corpus has a total of 625,065,294 characters and average character length of 6.07. IndicCorp has a total of 4,837,445,578 characters and average character length of 5.73, ∼5.93% less than that of Vācaspati. We performed a *t*-test to validate the null hypothesis that the average word length of a newspaper corpus like IndicCorp is always equal to or greater than that of a literary corpus like Vācaspati. The p-value from this test is less than 0.01, allowing us to reject the null hypothesis and conclude that the average word length of a literary corpus will consistently exceed that of a newspaper corpus. Languages with *rich morphology* tend to have longer words (e.g: Finnish, German, Bengali) whereas languages with *isolating grammar* (e.g., **Chinese, Vietnamese**) tend to have shorter words.Hence, Vācaspati, a literary corpus, is more adaptable to capture the rich morphology of Bangla compared to IndicCorp, the newspaper corpus. It is also observed that average word length ∼0.7% after merging Vācaspati with IndicCorp.

| Metric | Vācaspati | IndicCorp | Merged |
| --- | --- | --- | --- |
| Total Words | 102,900,420 | 844,141,105 | **947,041,525** |
| Total Sentences | 11,956,678 | 63,803,428 | **75,760,106** |
| Total Characters | 625,065,294 | 4,837,445,578 | **5,462,510,872** |
| Avg. Sentence Length (Words) | 8.61 | **13.23** | 12.50 |
| Avg. Word Length (Characters) | **6.07** | 5.73 | 5.77 |

**Table 1:** Lexical comparison of Vācaspati (Literary) IndicCorp (Newspaper) and merged corpus.

## 3.2 Unique Words and Type to Token Ratio (TTR)

One distinctive way of identifying lexical diversity in a corpus is by analyzing the number of unique words. **Unique words** refers to the number of distinct word types in a corpus, which is essential for assessing the richness of vocabulary. It influences tasks such as **language modelling, text generation, and corpus-based studies**. Vācaspati has 3,685,493 uniquewords whereas IndicCorp has 5,592,634 unique words. The absolute number of unique words is more for both IndicCorp, but they also encompass ∼4 and ∼8 times more words than Vācaspati, respectively. Thus, we further implore the number of unique words by randomly selecting 10M to 100M tokens from each corpora. We have chosen 10 subsamples for each token size, and the mean is reported in Table 2.

It is observed from Table 2 that Vācaspati (literary data) has more unique words for each sample of total words from 10M to 100M. We have fitted Heaps' law [10] (App A.1) on both IndicCorp and Vācaspati corpora to estimate the number of unique words each corpus will have at 844,141,105 words. Heaps' law predicts that IndicCorp will have 5,808,811 unique words at 844,141,105 total words which is within 5.44% of the actual, justifying Heaps' law's use in this case. Figure 1 illustrates the number of unique words for Vācaspati when matched to the total number of words in IndicCorp. The figure demonstrates that the number of unique words in a literary corpus will consistently be significantly higher than that in a newspaper corpus.

To further evaluate the lexical diversity of a corpus, we adopt the measure

| # Tokens | Vācaspati | IndicCorp |
|---|---|---|
| 10M | **802,984** | 394,727 |
| 20M | **1,281,517** | 604,456 |
| 30M | **1,682,468** | 772,457 |
| 40M | **2,033,927** | 918,932 |
| 50M | **2,358,358** | 1,052,008 |
| 60M | **2,656,294** | 1,174,913 |
| 70 M | **2,935,413** | 1,288,718 |
| 80M | **3,197,581** | 1,398,587 |
| 90M | **3448542** | 1,500,790 |
| 100M | **3,685,493** | 1,598,322 |

**Table 2:** Unique word comparison of Literary, Newspaper against different numbers of tokens.

**Type-Token Ratio (TTR)** [5] which compares the number of unique words to the total number of words in the corpus:

$$TTR = \frac{\text{Unique Words}}{\text{Total Words}} \times 100 \quad (2)$$

Vācaspati has a TTR of 3.53% where as TTR of IndicCorp is 0.65% (∼4.43 times smaller). A higher TTR indicates greater *vocabulary richness*, while a lower TTR suggests frequent repetition of words. To support the notion that literary corpora generally possess a higher TTR than newspaper corpora, we conducted *t*-testing. The p-value obtained is <0.01, indicating that the result is statistically significant.

Further imploration of the corpora revealed that there are 16.69% of non-Bangla tokens in IndicCorp corpora. In contrast, Vācaspati has only 1.93% of non-Bangla tokens, highlighting its native linguistic richness. To evaluate vocabulary richness, we further compute the Type-to-Token Ratio (TTR) of only Bangla tokens across varying chunk sizes from 1000 to 100M token sizes. TTR serves as an indicator of lexical diversity, with higher values indicating greater variation in word usage. Table 3 demonstrates the performance of TTR across various chunk sizes for all the corpora.

From Table 3 (Fig 2), it is observed that at smaller chunk sizes (e.g., 1,000 tokens), *IndicCorp* exhibit higher TTR than literature, but as the token size increases to 1M, the Vācaspati corpus outperforms IndicCorp in TTR. It reflects the lexical richness of Vācaspati (literary) corpora, highlighting the value of literature in providing a naturally diverse and expressive vocabulary.

### 3.2.1 Impact of Named Entities on Unique Words

The IndicCorp corpus exhibits a significantly higher number of unique words compared to Vācaspati, suggesting that the newspaper corpus is more lexically diverse than the literary corpus. In the previous section, we established that the literary corpus consistently has more unique words than the newspaper corpus. In this section, we tried to delve deeply into analysing the newspaper corpus, which has more unique words than the literary corpus. A large number of unique words can stem from a large number of named entities (N.Es). Newspaper typically encompasses wider range of N.Es like প্রেসিডেন্ট (prēsiḍēnṭa, President), সিইও (siiō, CEO) than a literary corpus.

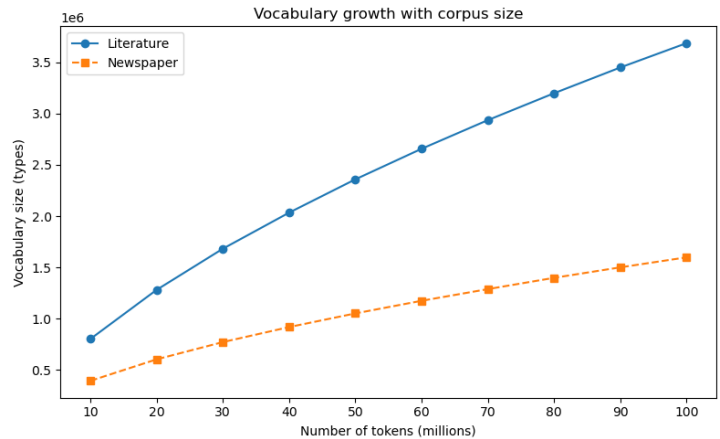

**Figure 1:** Figure showing expected number of unique words of VĀCASPATI when aligned with IndicCorp size.

Given the impracticality to label large corpus like VĀCASPATI (102M+ words) and IndicCorp (844M+ words), we decided to assess the quantity of proper nouns within each corpus, since proper nouns are inherently N.Es. To accomplish this, we used the publicly available iNLTK module [1] and used its Indian language part-of-speech (P.O.S) models for proper noun detection in Bangla. The module has an accuracy of 87.29%. To further validate the accuracy of iNLTK modules, we curated a ground truth dataset of 650 proper nouns (names of people, countries and others), which in turn are also N.Es. The accuracy of the iNLTK module is 88.79%, which is comparable to the reported accuracy, and thus, we can consider the resultant output from the iNLTK module to be reasonably good.

Our analysis show that VĀCASPATI has 197,436 proper nouns (∼0.19% of the words) which is significantly (∼1.21 times) lesser than IndicCorp comprising 3,575,085 proper nouns (∼0.42% words). We evaluated the result using *t*-testing, and the p-value <0.01 indicates that the result is statistically significant.

To further support our hypothesis that newspaper corpora contain significantly more named entities (N.Es) than literary corpora, we incorporated 100 additional N.Es, such as প্রেসিডেন্ট (prēsiḍēnṭa, President), প্রধানমন্ত্রী (pradhānamantrī, Prime Minister), মুখ্যমন্ত্রী (mukhyamantrī, Chief Minister), রাজ্যপাল (rājyapāla, Governor), নগরপাল (nagarapāla, Police Commissioner), সিইও (siiō, CEO), among others, to the existing 650 proper nouns resulting in a curated ground truth of 750 words. VĀCASPATI has a cumulative 589,941 (∼0.57%) occurrences of such words compared to 5,756,771 (∼1.27%) occurrences of those words in IndicCorp corpora which is ∼1.23 times lesser than IndicCorp. We again fitted Heaps' law to approximate the number of N.Es VĀCASPATI will have when both VĀCASPATI and IndicCorp have same number of words (∼844M). Fig 3 demonstrates the expected number of proper nouns for a literary corpus like VĀCASPATI aligned with the size of a newspaper corpus such as IndicCorp. The figure demonstrates that VĀCASPATI will consistently have fewer N.Es than IndicCorp.

| Tokens | Literature | Newspaper |
| --- | --- | --- |
| 1,000 | 63.08 ± 5.74 | **75.28 ±2.75** |
| 5,000 | 44.95 ± 5.85 | **57.40 ±1.20** |
| 10,000 | 37.87 ± 5.85 | **48.90 ±0.80** |
| 50,000 | 24.46 ± 5.44 | **30.65 ±0.29** |
| 100,000 | 20.13 ± 5.14 | **24.23 ±0.18** |
| 500,000 | 12.87 ± 4.27 | **13.30 ±0.06** |
| 1,000,000 | **10.67 ±3.92** | 10.12 ± 0.04 |
| 2,000,000 | **8.97 ±3.21** | 7.67 ± 0.02 |
| 3,000,000 | **8.06 ±2.79** | 6.51 ± 0.02 |
| 4,000,000 | **7.52 ±2.44** | 5.80 ± 0.02 |
| 5,000,000 | **7.16 ±1.99** | 5.30 ± 0.01 |
| 10,000,000 | **6.10 ±1.47** | 4.01 ± 0.01 |
| 15,000,000 | **5.68 ±1.04** | 3.40 ± 0.01 |
| 20,000,000 | **5.20 ±0.89** | 3.03 ± 0.01 |
| 30,000,000 | **4.87 ±0.32** | 2.57 ± 0.00 |
| 40,000,000 | **4.60 ±0.72** | 2.29 ± 0.00 |
| 50,000,000 | **4.20 ±0.67** | 2.10 ± 0.00 |
| 60,000,000 | **4.25 ±0.00** | 1.95 ± 0.00 |
| 70,000,000 | **4.16 ±0.00** | 1.83 ± 0.00 |
| 80,000,000 | **3.90 ±0.00** | 1.73 ± 0.00 |
| 90,000,000 | **3.68 ±0.00** | 1.65 ± 0.00 |
| 100,000,000 | **3.53 ±0.00** | 1.58 ± 0.00 |

**Table 3:** Mean and standard deviations of TTR for varying token chunks for all the corpora.

### 3.3 Hapax Legomena Ratio (HLR)

Hapax Legomena [18] are words that occur only once in a given text. The Hapax Legomena Ratio [18], defined as the proportion of hapax legomena to the total number of words, provides insights into a corpus's lexical diversity and richness.

$$HLR = \frac{\text{Hapax Legomena}}{\text{Total Words}} \quad (3)$$

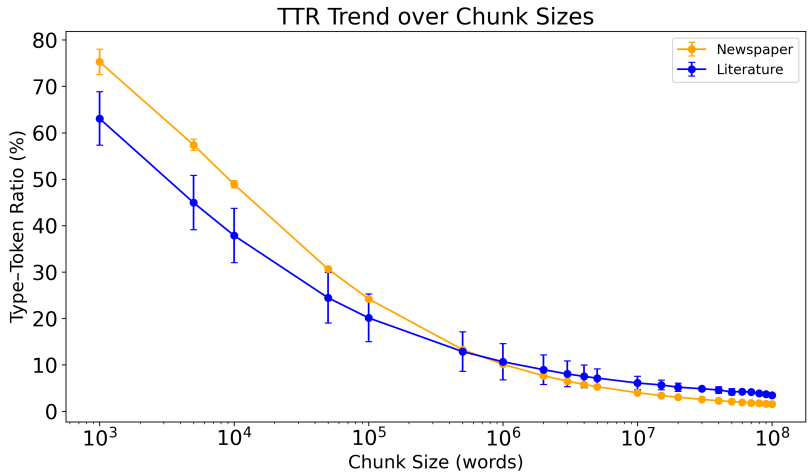

**Figure 2:** Figure showing the trend of TTR-chunks for both Vācaspati and Indic-Corp.

Hapax Legomena of Vācaspati is 2,000,000 with HLR of ∼ 1.72% whereas hapax legomena of IndicCorp is 3,000,000 with HLR of ∼0.36%. It establishes the lexical diversity and richness of a literary corpus.

We further evaluated how the HLR trend varies with different fractions of both Vācaspati and IndicCorp corpora, starting with 10% to 100% of each corpus. We have randomly selected 10 subsamples of each fraction and reported the mean and standard deviation (rounded off two decimal points) in Table 4.

| Fraction of corpus | Literature | Newspaper |
|---|---|---|
| 0.1 | **4.08 ±0.00** | 1.02 ± 0.00 |
| 0.2 | **3.22 ±0.00** | 0.76 ± 0.00 |
| 0.3 | **2.78 ±0.00** | 0.63 ± 0.00 |
| 0.4 | **2.51 ±0.00** | 0.55 ± 0.00 |
| 0.5 | **2.31 ±0.00** | 0.50 ± 0.00 |
| 0.6 | **2.15 ±0.00** | 0.46 ± 0.00 |
| 0.7 | **2.02 ±0.00** | 0.42 ± 0.00 |
| 0.8 | **1.91 ±0.00** | 0.40 ± 0.00 |
| 0.9 | **1.81 ±0.00** | 0.38 ± 0.00 |
| 1.0 | **1.72 ±0.00** | 0.36 ± 0.00 |

**Table 4:** Variation of HLR fraction of corpus for literary and newspaper data.

Table 4 and Figure 4 shows that the HLR monotonically decreases for both newspaper and literature corpus. However HLR of Vācaspati is always higher than that of IndicCorp. The rate of decrease of HLR of Vācaspati is lower than that IndicCorp. From fraction 0.1 to 0.2 the rate of decrease for VĀCASPATI is 21.07% whereas that of IndicCorp is 25.49% (∼21% higher than VĀCASPATI). Similarly from fraction size 0.9 to 1.0 the rate of decrease of HLR of VĀCASPATI (4.97%) is ∼5.51% is less than that of IndicCorp is (5.26%). These findings indicate the lexical richness of VĀCASPATI.

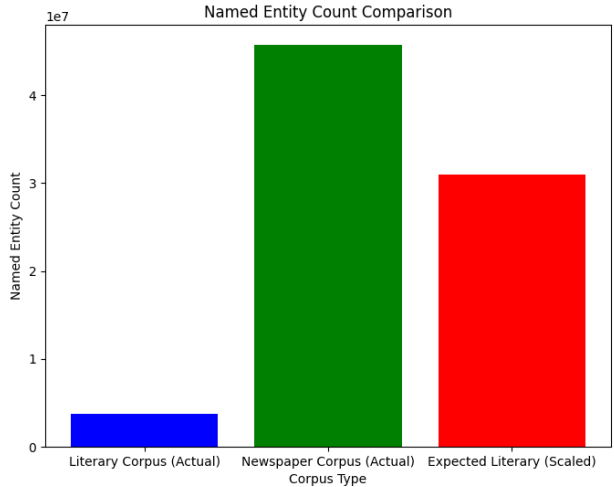

**Figure 3:** Figure showing expected number of 750 named entities of VĀCASPATI when aligned with IndicCorp size.

### 3.4 Bigram Diversity

Bigram diversity serves as a key linguistic metric to analyze the structural complexity of a corpus. It quantifies the variety of two-word sequences (bigrams) and helps assess the complexity of textual structures. It is computed using the following formula:

$$\text{Bigram Diversity} = \frac{\text{Unique Bigrams}}{\text{Total Bigrams}} \quad (4)$$

where:

- **Unique Bigrams**: The count of distinct two-word sequences in the text.

- **Total Bigrams**: The total number of bigrams in the corpus.

A higher bigram diversity score indicates a more varied and complex text, while a lower score suggests repetition. VĀCASPATI has a bigram diversity of 47.07%, ∼76% higher than IndicCorp, which has a bigram diversity of 26.75%, suggesting a richer and more expressive vocabulary for a literary corpus.

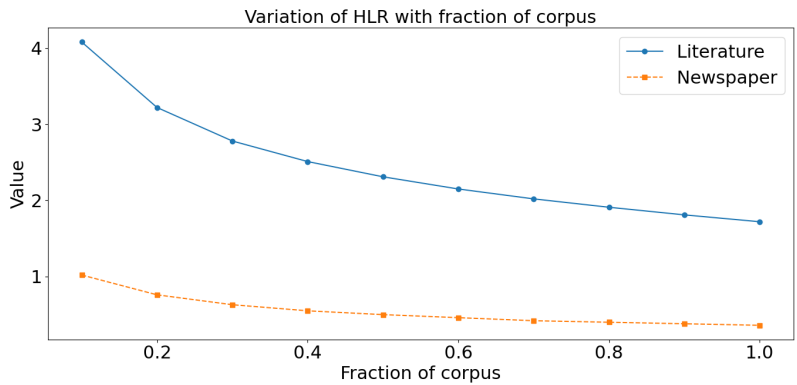

**Figure 4:** Figure showing the hapax legomena ratio of Vācaspati and IndicCorp corpus at different fractions of their total size.

We further evaluated how the bigram diversity varies with different fractions of both Vācaspati and IndicCorp corpora, starting with 10% to 100% of each corpus. We have randomly selected 10 subsamples of each fraction and reported the mean and standard deviation (rounded off two decimal points) in Table 5.

| **Fraction of corpus** | **Literature** | **Newspaper** |
|---|---:|---:|
| 0.1 | **64.93 ±0.00** | 43.53 ± 0.00 |
| 0.2 | **59.69 ±0.00** | 38.09 ± 0.00 |
| 0.3 | **56.69 ±0.00** | 35.03 ± 0.00 |
| 0.4 | **54.37 ±0.00** | 32.95 ± 0.00 |
| 0.5 | **52.66 ±0.00** | 31.37 ± 0.00 |
| 0.6 | **51.27 ±0.00** | 30.11 ± 0.00 |
| 0.7 | **50.10 ±0.00** | 29.07 ± 0.00 |
| 0.8 | **49.09 ±0.00** | 28.19 ± 0.00 |
| 0.9 | **48.20 ±0.00** | 27.30 ± 0.00 |
| 1.0 | **47.07 ±0.00** | 26.75 ± 0.00 |

**Table 5:** Variation of bigram diversity with fraction of corpus for literary, and newspaper.

Table 5 and Figure 5 shows that the bigram diversity monotonically decreases for both newspaper and literature corpus. However HLR of Vācaspati is always higher than that of IndicCorp. The rate of decrease of HLR of Vācaspati is lower than that IndicCorp. From fraction 0.1 to 0.2 the rate of decrease for Vācaspati is 8.07% whereas that of IndicCorp is 12.50.49% (∼35.44% higher than Vācaspati). These findings indicate the lexical richness of Vācaspati.

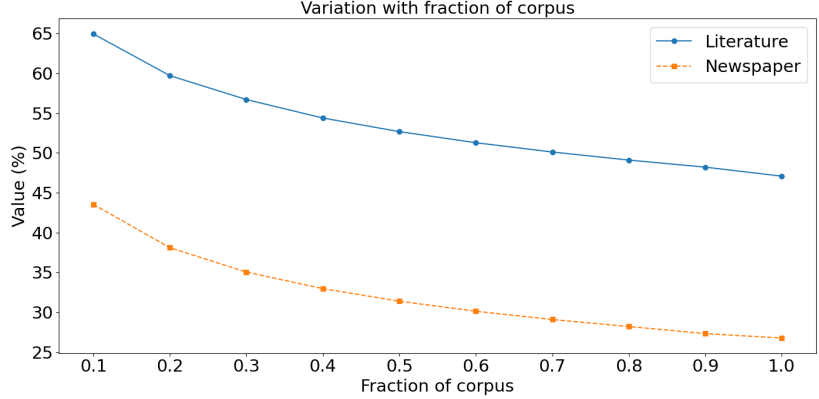

**Figure 5:** Figure showing the bigram diversity of Vācaspati, IndicCorp and Merged corpus at different fractions of their total size.

## 3.5 Average Syllable Length

Another important metric for understanding the structure and richness of a corpus is the average syllable count for a corpus. This section will investigate these measures for Vācaspati and IndicCorp corpus. Bangla syllables are determined by the presence of vowels and vowel signs within a word. We used the following approach to compute the syllable count in Bangla text:

- **Independent vowels:** We count segments of words ending with following vowels অ, আ, ই, ঈ, উ, ঊ, ঋ, এ, ঐ, ও, ঔ ( a, ā, i, ī, u, ū, r̥, ē, ai, ō, au) as one syllable.

- **Vowel signs modifying consonants:** We have also counted segment of words ending with following modifiers া, ি, ী, ু, ূ, ৃ, ে, ৈ, ো, ৌ ( ā, i, ī, u, ū, r̥, ē, ai, ō, au) as one syllable.

Vācaspati has a total of 192,627,558 syllables whereas IndicCorp has 1,287,263,766 syllables. The lexical diversity of a corpus is identified by *average syllable length* of the corpus.

$$\text{Avg Syllables per Word} = \frac{\text{Total Syllables}}{\text{Total Words}} \quad (5)$$

Vācaspati has average syllable length of 1.72 whereas IndicCorp has average syllable length of 1.52. A higher value indicates a lexically dense and complex text, while a lower value suggests simpler language usage. It indicates that a literary corpus like Vācaspati will be more lexically diverse than a newspaper corpus like IndicCorp.

Vācaspati outperforms IndicCorp in all of TTR, HLR, Bigram Diversity and Avg Syllable length, indicating its lexical richness. Analysis of subsamples of fractions of Vācaspati and IndicCorp also validates this hypothesis.

Appendix A.2 and Appendix A.3 also shows that performance of merged corpora is better than that of IndicCorp indicating that addition of literary data increases the lexical richness of corpora.

# 4 Zipf's Law

Zipf's Law is a well-known empirical rule describing word frequency distribution in natural languages. It states that the frequency of any word is inversely proportional to its rank in the frequency table. If $f$ is the frequency of a word and $r$ is its rank, then Zipf's Law is expressed as:

$$f(r) \propto \frac{1}{r}$$

Zipf's law implies that the most frequent word occurs approximately twice as often as the second most frequent word, three times as often as the third most frequent word, and so on. When plotted on a log-log scale, the rank-frequency distribution of words in a language typically forms a straight line, indicating a power-law distribution.

## 4.1 Rationale for Zipf's Law

Applying Zipf's Law to corpora serves several purposes:

- **Linguistic Insights:** Understanding the frequency distribution of words helps uncover a language's underlying structure and regularities.

- **Corpus Analysis:** It provides a method to compare different types of corpora, such as literary texts and newspaper articles, by analyzing their adherence to the power-law distribution. It aids in identifying the core vocabulary of a language, which is essential for tasks like language modelling and dictionary creation. Additionally, we can find the words causing the disbalance in a power-law distribution. These words are more pronounced; hence, we can create embeddings where additional weightage can be given to these words, which can help in Natural Language Processing (N.L.P).

- **Model Evaluation:** In computational linguistics, ensuring that models follow Zipfian distribution helps validate their performance and naturalness.

## 4.2 Zipf's Law for different Bangla Corpora

In this section, we analyzed three types of Bangla corpora: Vācaspati, IndicCorp, and Indic-Vācaspati corpora. We calculated the slopes of their rank-frequency distributions to understand how closely they adhere to Zipf's Law.

### 4.2.1 Vācaspati

The slope of the rank-frequency distribution for VĀCASPATI was −1.10. It indicates a distribution relatively close to the theoretical slope of −1.00. This suggests that the VĀCASPATI, predominantly a literary corpus, closely follows Zipf's Law, implying a well-balanced use of high-frequency and low-frequency words.

We further investigated the words causing the slight deviation from the power-law for VĀCASPATI. The words causing this deviation for VĀCASPATI are primarily stop-words in Bangla, like না (nā, no), করে (karē, do), তার (tāra, their), আমি (āmi, I) and others such words. অনেক (anēka, many) (rank: 81) and হাত (hāta, hand) (rank: 85) are few words which are not stopwords but Appendix A.5.

### 4.2.2 IndicCorp

Similar to the literary corpus VĀCASPATI, we found out the slope of the newspaper corpus IndicCorp, which was -1.36. This deviation from the ideal slope of -1 suggests a steeper distribution, meaning that high-frequency words are used more frequently compared to lower-frequency words than in the literary corpus. This could be due to the repetitive nature of news articles, which often use a more limited vocabulary to report on current events consistently.

The slope of the newspaper corpus was -1.36. This deviation from the ideal slope of -1 suggests a steeper distribution, meaning high-frequency words are used more frequently than lower-frequency words. News articles are often repetitive and use a limited vocabulary to report current events which can be the primary reason for such a steeper deviation. Presence of words like পুলিশ (puliśa, police) (rank: 58), বাংলাদেশ (bāṁlādēśa, Bangladesh) (rank: 78), দেশের (dēśēra, country's) (rank: 96), টাকা (ṭākā, money), and অভিযোগ (abhiyōga, accusation) (rank: 99) in the top 100 deviation list indicates in this direction only. The entire list of the top 100 words is shown in Appendix A.5.

On analysing the results of Zipf's law, we observe that the literary corpus has a slope closer to -1, the ideal slope. It indicates that the VĀCASPATI has a richer and more diverse vocabulary. In contrast, the newspaper corpus's steeper slope suggests a more repetitive vocabulary.

We further explored the effect of adding a fraction of the literary corpus to the full newspaper corpus and the other way round with respect to adherence to Zipf's law. We randomly subsampled 10 fractions of each category, ranging from 0.1 to 0.9 of the literature corpus, and merged them with the newspaper corpus. In the second case, we subsampled 10 fractions from each category of the newspaper corpus and merged them with the literature corpus. Table 6 records the mean and standard deviation of each fraction. It is evident from the table that the slope deviates more from the ideal value of -1 on the addition of a fraction of newspaper data to the complete literary data. On the other hand, the slope approaches the ideal value by adding a fraction of the literary data to the complete newspaper data. These results indicate that adding literary data enhances the lexical richness of the corpora, whereas adding newspaper data decreases it. Figure 6 shows the trend of zipf's law on adding a fraction of one corpus to the full of another.

| Literature Fraction | Newspaper Fraction | Mean |
|---:|---:|---:|
| 1.0 | 0.0 | $-1.1137 \pm 0.00$ |
| 1.0 | 0.1 | $-1.1630 \pm 0.00$ |
| 1.0 | 0.2 | $-1.1950 \pm 0.00$ |
| 1.0 | 0.3 | $-1.2190 \pm 0.00$ |
| 1.0 | 0.4 | $-1.2377 \pm 0.00$ |
| 1.0 | 0.5 | $-1.2530 \pm 0.00$ |
| 1.0 | 0.6 | $-1.2660 \pm 0.00$ |
| 1.0 | 0.7 | $-1.2780 \pm 0.00$ |
| 1.0 | 0.8 | $-1.2880 \pm 0.00$ |
| 1.0 | 0.9 | $-1.2967 \pm 0.00$ |
| 1.0 | 1.0 | $-1.3048 \pm 0.00$ |
| 0.0 | 1.0 | $-1.3660 \pm 0.00$ |
| 0.1 | 1.0 | $-1.3510 \pm 0.00$ |
| 0.2 | 1.0 | $-1.3410 \pm 0.00$ |
| 0.3 | 1.0 | $-1.3340 \pm 0.00$ |
| 0.4 | 1.0 | $-1.3282 \pm 0.00$ |
| 0.5 | 1.0 | $-1.3229 \pm 0.00$ |
| 0.6 | 1.0 | $-1.3185 \pm 0.00$ |
| 0.7 | 1.0 | $-1.3145 \pm 0.00$ |
| 0.8 | 1.0 | $-1.3109 \pm 0.00$ |
| 0.9 | 1.0 | $-1.3077 \pm 0.00$ |
| 1.0 | 1.0 | $-1.3048 \pm 0.00$ |

**Table 6:** Mean (and standard deviation) of the measure across different mixtures of literature and newspaper corpora.

## 5 Entropy and Redundancy

In this section we will investigate the entropy and redundancy pertaining to VĀCASPATI and IndicCorp.

### 5.1 Entropy

In Information theory **entropy** [21] measures the unpredictability in a probability distribution, where higher entropy means outcomes are more varied and less predictable. Entropy is defined as

$$\mathrm{H}(X) := -\sum_{x \in \mathcal{X}} p(x) \, \log p(x),$$

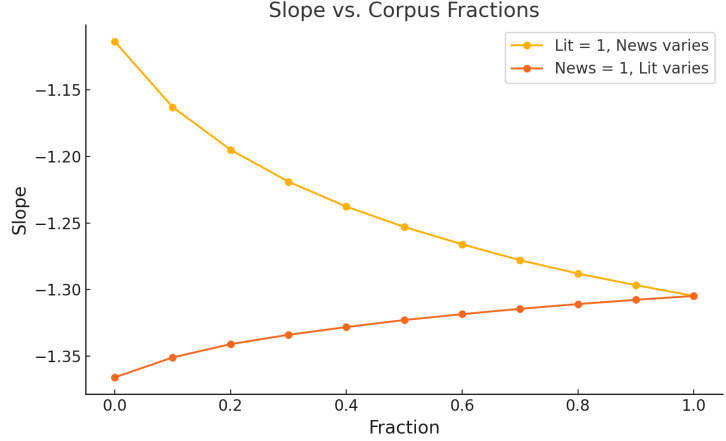

**Figure 6:** Figure showing zipf's law trends on adding fraction of one corpus to the full of another.

, where $x$ denotes a single possible outcome (event) of the random variable $X$, and $\mathcal{X}$ is the set of all such potential outcomes.

In computational linguistics, entropy [11] quantifies the unpredictability of word sequences in a corpus. For bigrams, it is computed as:

$$H_2 = -\sum_{w \in \mathcal{V}_2} P(w) \log_2 P(w)$$

, where $[\mathcal{V}_2]$ the set of all distinct bigrams (pairs of consecutive words) in the corpus and $[P(w)]$ the empirical probability of bigram $w = (w_i, w_{i+1})$. $[P(w)]$ is defined as:

$$P(w) = \frac{\text{count}(w_i, w_{i+1})}{\sum_{u \in \mathcal{V}_2} \text{count}(u)}.$$

Higher n-gram entropy indicates more diverse word patterns in the corpus, which makes the model uncertain about its predictions.

## 5.2 Redundancy

In information theory, redundancy [21] refers to the portion of a message that is structurally determined by statistical constraints, rather than conveying new information. It is the difference between the maximum possible information content for a given alphabet and the actual information content, that is, the entropy. For a discrete random variable $X$ with outcome set $\mathcal{X}$, redundancy is defined as

$$\text{R}(X) := 1 - \frac{\text{H}(X)}{\log_2 |\mathcal{X}|},$$

where $\text{H}(X)$ is the actual entropy of $X$ and $\log_2 |\mathcal{X}|$ is the maximum entropy.

In computational linguistics, redundancy [11] measures how much of the vocabulary potential is unused due to *repetition* in a corpus. For bigrams, we take $\mathcal{X} = \mathcal{V}_2$, the set of all distinct bigrams, and write

$$R_2 := 1 - \frac{H_2}{\log_2|\mathcal{V}_2|},$$

where $H_2$ is the bigram entropy

$$H_2 := -\sum_{w \in \mathcal{V}_2} P(w) \log_2 P(w),$$

and $P(w)$ is the empirical probability of bigram $w$. Higher redundancy indicates more repetitive and formulaic local word patterns, while lower redundancy indicates more information-rich corpora.

|  | Entropy | | | Redundancy (%) | | |
|---|---|---|---|---|---|---|
| **Type** | **Literary** | **Newspaper** | **Merged** | **Literary** | **Newspaper** | **Merged** |
| Unigram | 14.07 | 12.44 | 12.54 | 35.56 | 44.43 | 44.89 |
| Bigram | 22.56 | 20.91 | 21.10 | 10.30 | 20.92 | 20.74 |
| Trigram | 25.67 | 25.70 | 25.90 | 2.04 | 8.73 | 8.43 |

**Table 7:** Entropy and redundancy for VĀCASPATI, IndicCorp, and merged corpora.

Table 7 and shows the entropy and redundancy values of VĀCASPATI, IndicCorp and merged corpora for unigram, bigram and trigram. From Table 7 VĀCASPATI has higher entropy than IndicCorp and merged for unigrams and bigrams, whereas in trigrams, all three corpora have similar entropy. However, VĀCASPATI has a significantly lower redundancy value compared to IndicCorp and Merged dataset for all n-grams. This indicates that VĀCASPATI has a more information-rich corpus.

However, these results are for the entire corpus of both VĀCASPATI and IndicCorp, which contains different numbers of sentences. To validate our assumption that literary corpora are more information-rich, we subsampled an equal number of sentences from both VĀCASPATI and IndicCorp datasets. Specifically, we selected 10 random subsamples from a group of sentences, and the mean and standard deviation (rounded to 2 digits) are reported.

Table 8 and Table 9 show the entropy and redundancy scores of VĀCASPATI and IndicCorp as a function of the number of sentences. From Table 8 and Table 9 it is observed that the entropy value increases with an increase in the number of sentences for both IndicCorp and VĀCASPATI. However, the entropy of VĀCASPATI is always higher than that of IndicCorp. Table 8 and Table 9 also demonstrate that the redundancy value of both VĀCASPATI and IndicCorp increases with an increase in the number of sentences. However, VĀCASPATI has significantly lower redundancy than IndicCorp. The results indicate that literary datasets, such as VĀCASPATI, are more information-rich than newspaper datasets, like IndicCorp.

|  | Entropy | | | Redundancy (%) | | |
|---:|---|---|---|---|---|---|
| #Sentences | Unigram | Bigram | Trigram | Unigram | Bigram | Trigram |
| 1,000 | $12.07 \pm 0.06$ | $14.34 \pm 0.01$ | $14.44 \pm 0.01$ | $9.00 \pm 0.00$ | $0.20 \pm 0.00$ | $0.02 \pm 0.00$ |
| 5,000 | $12.89 \pm 0.03$ | $16.48 \pm 0.02$ | $16.76 \pm 0.02$ | $14.00 \pm 0.00$ | $0.70 \pm 0.00$ | $0.04 \pm 0.00$ |
| 10,000 | $13.15 \pm 0.03$ | $17.34 \pm 0.02$ | $17.75 \pm 0.02$ | $16.00 \pm 0.00$ | $1.20 \pm 0.00$ | $0.06 \pm 0.00$ |
| 50,000 | $13.58 \pm 0.01$ | $19.15 \pm 0.01$ | $20.02 \pm 0.01$ | $22.00 \pm 0.00$ | $2.60 \pm 0.00$ | $0.16 \pm 0.00$ |
| 100,000 | $13.71 \pm 0.01$ | $19.85 \pm 0.00$ | $20.98 \pm 0.00$ | $24.00 \pm 0.00$ | $3.40 \pm 0.00$ | $0.25 \pm 0.00$ |
| 250,000 | $13.84 \pm 0.01$ | $20.68 \pm 0.00$ | $22.21 \pm 0.00$ | $27.00 \pm 0.00$ | $4.80 \pm 0.00$ | $0.43 \pm 0.00$ |
| 500,000 | $13.92 \pm 0.01$ | $21.24 \pm 0.00$ | $23.12 \pm 0.00$ | $29.00 \pm 0.00$ | $5.93 \pm 0.00$ | $0.63 \pm 0.00$ |
| 1,000,000 | $13.98 \pm 0.00$ | $21.73 \pm 0.00$ | $24.00 \pm 0.00$ | $31.00 \pm 0.00$ | $7.18 \pm 0.00$ | $0.90 \pm 0.00$ |
| 2,000,000 | $14.03 \pm 0.002$ | $22.39 \pm 0.00$ | $24.82 \pm 0.00$ | $33.00 \pm 0.00$ | $8.50 \pm 0.00$ | $1.26 \pm 0.00$ |

**Table 8:** Entropy and redundancy as a function of the number of sentences for Vācaspati.

|  | Entropy | | | Redundancy (%) | | |
|---:|---|---|---|---|---|---|
| #Sentences | Unigram | Bigram | Trigram | Unigram | Bigram | Trigram |
| 1,000 | $11.09 \pm 0.02$ | $13.95 \pm 0.06$ | $14.31 \pm 0.07$ | $14.15 \pm 0.00$ | $1.56 \pm 0.00$ | $0.20 \pm 0.00$ |
| 5,000 | $11.71 \pm 0.01$ | $15.85 \pm 0.03$ | $16.56 \pm 0.00$ | $19.25 \pm 0.00$ | $2.97 \pm 0.00$ | $0.45 \pm 0.00$ |
| 10,000 | $11.89 \pm 0.01$ | $16.57 \pm 0.01$ | $17.49 \pm 0.02$ | $21.53 \pm 0.00$ | $3.79 \pm 0.00$ | $0.60 \pm 0.00$ |
| 50,000 | $12.16 \pm 0.00$ | $18.02 \pm 0.01$ | $19.58 \pm 0.01$ | $26.74 \pm 0.00$ | $6.30 \pm 0.00$ | $1.24 \pm 0.00$ |
| 100,000 | $12.23 \pm 0.00$ | $18.54 \pm 0.01$ | $20.44 \pm 0.01$ | $28.91 \pm 0.00$ | $7.61 \pm 0.00$ | $1.64 \pm 0.00$ |
| 250,000 | $12.30 \pm 0.00$ | $19.13 \pm 0.01$ | $21.51 \pm 0.01$ | $31.68 \pm 0.00$ | $9.52 \pm 0.00$ | $2.31 \pm 0.00$ |
| 500,000 | $12.34 \pm 0.00$ | $19.50 \pm 0.00$ | $22.26 \pm 0.00$ | $33.70 \pm 0.00$ | $11.07 \pm 0.00$ | $2.92 \pm 0.00$ |
| 1,000,000 | $12.37 \pm 0.00$ | $19.83 \pm 0.00$ | $22.97 \pm 0.00$ | $37.51 \pm 0.00$ | $14.36 \pm 0.00$ | $4.45 \pm 0.00$ |
| 2,000,000 | $12.39 \pm 0.00$ | $20.10 \pm 0.00$ | $23.63 \pm 0.00$ | $38.56 \pm 0.00$ | $15.34 \pm 0.00$ | $4.97 \pm 0.00$ |

**Table 9:** Entropy and redundancy as a function of the number of sentences for IndicCorp.

# 6 Perplexity

Perplexity [11] is a key metric that quantifies a language model's uncertainty when predicting a sequence of words. For a sequence $w_1^N = (w_1, w_2, \ldots, w_N)$, the perplexity of a model $P$ is defined as

$$\text{PP}(w_1^N) := P(w_1, w_2, \ldots, w_N)^{-\frac{1}{N}}.$$

Perplexity is usually expressed in terms of the average log-probability:

$$\text{PP}(w_1^N) = \exp\left(-\frac{1}{N} \sum_{i=1}^{N} \log P(w_i \mid w_1^{i-1})\right).$$

It can be further refined as

$$\text{PP}(w_1^N) = 2^{-\frac{1}{N} \sum_{i=1}^{N} \log_2 P(w_i \mid w_1^{i-1})}.$$

Here, $w_1^N$ denotes the sequence $(w_1, w_2, \ldots, w_N)$, and $P(w_i \mid w_1^{i-1})$ is the model's probability of the $i$-th token $w_i$ given $w_1^{i-1}$.

Lower perplexity indicates that the model is more confident in its predictions and can understand the patterns in the corpora. Higher perplexity, on the other hand, suggests that the model is uncertain about its prediction. It can be either due to poor training or the corpus is lexically diverse, and thus the model is unable to recognise patterns.

## 6.1 Perplexity analysis of Vācaspati

To evaluate the predictability of different corpora, we trained word-based $n$-gram language models (for $n = 1, 2, 3$) on both the IndicCorp and VĀCASPATI. The word-based $n$-gram model is a simple model that learns directly from the statistical patterns present in the corpus. Deep learning architectures, on the other hand, also depend on loss functions and optimizers for understanding a language. Since we are trying to understand the predictability of a corpus, we focused on $n$-gram modelling with perplexity as the evaluation metric. From both the IndicCorp and VĀCASPATI, we at first randomly subsample 5 non-overlapping corpora of ∼4M sentences. We build separate unigram, bigram and trigram models for each of these corpora. We employed a 10-fold cross-validation on each of these corpora, and the mean and standard deviation for each of the $n$-gram models are reported in Table 10.

Across all three $n$-gram orders, we consistently observe that the Bangla newspaper corpus yields lower perplexity than the Bangla literature corpus. Since both IndicCorp and VĀCASPATI are high-quality corpora, and our methodology involved only building $n$-gram models, the high perplexity value can be attributed to the diversity of these corpora. VĀCASPATI is more diverse compared to IndicCorp with richer vocabulary, and more varied syntactic structures. In contrast, IndicCorp has a higher number of named entities and a set of frequently recurring constructions. Thus, the perplexity of $n$-gram models built on VĀCASPATI is higher than that of IndicCorp dataset.

|  | Literature | | | Newspaper | | |
|---|---|---|---|---|---|---|
| **Language** | **Unigram** | **Bigram** | **Trigram** | **Unigram** | **Bigram** | **Trigram** |
| Bangla | $8423.27 \pm 1.01$ | $2003.89 \pm 1.24$ | $13430.35 \pm 39.38$ | $5229.79 \pm 6.30$ | $881.20 \pm 1.56$ | $6347.71 \pm 28.76$ |
| English | $2614.22 \pm 0.61$ | $959.87 \pm 1.01$ | $8363.07 \pm 14.03$ | $1895.78 \pm 2.56$ | $390.60 \pm 8.18$ | $1654.12 \pm 2.05$ |

**Table 10:** Perplexity of $n$-gram models for Bangla and English literature and newspaper corpora.

To examine whether this pattern is specific to Bangla or reflects a broader phenomenon, we repeated the same experimental setup on English newspapers and English literature corpora. We downloaded the literary dataset associated with the **Gutenberg Project** (https://www.gutenberg.org/) and newspaper data between 2007 to 2024 from the newspaper data-crawl (https://data.statmt.org/news-crawl/en/) website. We followed the same setting as employed for Bangla by creating 5 subsamples of ∼4M sentences. Each subsampled then undergoes 10-fold cross-validation. Mean and standard deviation of this process are reported in Table 10. From the Table 10 (Figure 13 of App A.7) , it is observed that like Bangla, English literary data have higher perplexity than English newspaper data. English literary data also has higher perplexity than Bangla newspaper data. This suggests that the higher perplexity of literature is not specific to Bangla data, instead is associated with the greater lexical and structural diversity of literary writing. A second trend emerges when we compare Bangla and English within the same domain. For both literature and newspaper, English corpora exhibit lower perplexity than their Bangla counterparts at all $n$-gram orders. This result indicates that, at similar corpora size, simple $n$-gram models find Bangla surface forms harder to predict than English surface forms, indicating that the modelling techniques employed for English may not be efficient for Bangla.

# 7 Readability of Bangla Literature and Newspaper Corpus

Readability assessment is essential in evaluating the complexity and accessibility of written text and corpora. Several established readability metrics have been used in computational linguistics to measure text difficulty. Some of the popularly used readability metric are *Flesch Reading Ease (FRE)* [7], *Flesch-Kincaid Grade Level (FKGL)* [14], and Coleman-Liau Index (CLI) [4]. These metrics were originally developed for English but can be applied to other languages, including Bangla, without modification. Since they rely on structural properties such as sentence length, word length, and syllables per word, they remain effective across languages with similar writing systems.

## 7.1 Flesch Reading Ease (FRE)

The **Flesch Reading Ease Score (FRE)** [7], measures readability based on the following formula:

$$FRE = 206.835 - 1.015 \times \frac{\text{total words}}{\text{total sentences}} - 84.6 \times \frac{\text{total syllables}}{\text{total words}} \quad (6)$$

The lower the FRE score, the more complex the texts are and will require advanced language knowledge for readability. Table 11 shows the readability score and the education level required for reading the text. This interpretation is for English; however, Bangla, like English, uses syllable-based word formation. It allows us to adapt the FRE formula directly without modifications.

13 presents the FRE score of IndicCorp and VĀCASPATI corpus. The FRE score of IndicCorp is closer to *Reader's digest* magazine (FRE∼65), and 8-9th class students can read it competently. FRE score of VĀCASPATI is 52.59, similar to *Time* magazine (FRE∼52), and even for 12th-grade students, it is fairly difficult to read.

## 7.2 Coleman-Liau Index (CLI)

The Coleman-Liau Index (CLI) is a modern technique designed to gauge the understandability of a text. It uses character-based measures instead of syllables:

$$CLI = 0.0588 \times L - 0.296 \times S - 15.8 \quad (7)$$

| Score | School Level (US) |
|---|---|
| 100.00-90.00 | 5th grade |
| 90.0-80.0 | 6th grade |
| 80.0-70.0 | 7th grade |
| 70.0-60.0 | 8th & 9th grade |
| 60.0-50.0 | 10th to 12th grade |
| 50.0-30.0 | College |
| 30.0-10.0 | College graduate |
| 10.0-0.0 | Professional |

**Table 11:** Flesch Reading Ease Score Interpretation

where:

- $L$ = Average number of letters per 100 words.
- $S$ = Average number of sentences per 100 words.

| Score | School Level (US) | Age Range |
|---|---|---|
| 0–1 | Pre-K - 1st grade | 3-7 |
| 1–5 | 1st grade - 5th grade | 7-11 |
| 5–8 | 5th grade - 8th grade | 11-14 |
| 8–11 | 8th grade - 11th grade | 14-17 |
| 11 and above | 11th grade - college | 17 and above |

**Table 12:** Readability Levels and Age Interpretation

A grade level of 14.5 is considered understandable by a second-year undergraduate in the American Education system. Both IndicCorp and VĀCASPATI have Coleman Liau index higher than 14.5, making them fairly difficult to understand. VĀCASPATI also has a higher score (Table 13) than IndicCorp, making literary corpus more challenging to understand than a newspaper corpus. The merged corpus of a newspaper and literature can serve as a middle ground as observed from Table 13.

Understanding readability differences is crucial for various NLP applications, help in designing Bangla learning materials (*Learners' corpora*) for different age groups. VĀCASPATI can help in that because of its high FRE score.

| Corpus     | Flesch Reading Ease | Coleman-Liau Index |
|------------|---------------------|--------------------|
| Newspaper  | 64.81               | 15.65              |
| Literature | 52.59               | 16.43              |
| Merged     | 63.4                | 15.76              |

**Table 13:** Readability Scores of Bangla Texts

# 8 Effect of Vācaspati on Downstream Tasks

In this section, we examine the impact of VĀCASPATI on various downstream tasks. We assessed the effect of VĀCASPATI in two ways. In the first case, we tried to detect and separate merged words in a sentence through the lexicon obtained from VĀCASPATI. In the other case, we trained Electra models with by merging fractions of VĀCASPATI and IndicCorp dataset and compared their performance on various downstream tasks. In this section, we will discuss these two ways in detail.

## 8.1 Real-Word Merge Detection

Real-word merge is an error where two valid words are incorrectly joined or merged into one word. Real-world merge detection focuses on detecting these invalid merges and then generating the corresponding correct sentence. For example, for the sentence "উত্তম একজন অসাধারণঅভিনেতা।" ( uttama ēkajana asādhāraṇaabhinētā.) this task will detect the wrong merge অসাধারণঅভিনেতা and will generate the correct sentence "উত্তম একজন অসাধারণ অভিনেতা।" ( uttama ēkajana asādhāraṇa abhinētā., Uttam is a great actor).

Our objective in this task is to test the impact of lexicons generated from a corpus in detecting and correcting the invalid merges. For this task, we first created a test set of 20,000 Bangla sentences by prompting GPT-4o. The sentences generated are between 5 and 30 words. 17,500 sentences of the test set contain only one invalid merge, whereas 2,500 sentences consist of 2 to 5 invalid merges.

## 8.2 Methodology

We first build a vocabulary from a large, clean corpus. This set is treated as the list of "valid words" in the language. We then employ a greedy-based approach to detect and correct merges for each test sentence.

- If the token is already in the vocabulary $V$, it is kept as is.

- Otherwise, we try to split the token into exactly two parts that are both valid words in $V$.

- The splitting is done greedily from right to left: we first try the longest possible left part (i.e., splitting near the end), then move the split point leftwards until we find a pair (left, right) where both substrings belong to $V$.

- If such a split is found, those two words replace the original token. If no valid bipartition exists, the token is left unchanged.

We run this experiment by building vocabulary from VĀCASPATI, IndicCorp and Merged separately. Table 14 shows the performance of this method after building vocabulary from each dataset. We used GLEU and Accuracy as evaluation metrics for the task. From the Table, it is observed that the method with vocabulary created from VĀCASPATI outperforms all others, both for single-merged and multiple-merged sentences. The performance of this greedy method with the Merged dataset is better than that with only the IndicCorp dataset. Since no deep-learning techniques are involved, this performance can be attributed solely to the effect of the vocabulary created from the dataset. This result highlights the quality and effectiveness of VĀCASPATI for downstream tasks.

|  | Single Error | | Multi Error | | Overall | |
|---|---|---|---|---|---|---|
| Corpus | GLEU | Acc | GLEU | Acc | GLEU | Acc |
| Literature | **94.85%** | **91.89%** | **27.72%** | **16.16%** | **86.47%** | **82.42%** |
| Newspaper | 86.73 | 78.87% | 24.00 | 11.20% | 78.89 | 70.42% |
| Merged | 87.88 | 80.83% | 24.67 | 12.04% | 79.99 | 72.23% |

**Table 14:** Word Boundary Detection Performance across Corpora

## 8.3 Transformer Models

In this section, we aim to assess the impact of VĀCASPATI on downstream tasks after pretraining an Electra-small ($sim$16M parameter) model. We focused on the downstream tasks *Poem Classification*, *Sentiment Classification*, *Spelling Error Detection* and *NLI*. We trained electra small model by merging $\frac{1}{3}$ and $\frac{1}{2}$ of VĀCASPATI with full of IndicCorp dataset separately. Similarly we trained electra-small models with $\frac{1}{3}$ and $\frac{1}{2}$ of IndicCorp dataset with full of VĀCASPATI. We then evaluated their performance on the aforementioned downstream tasks.

### 8.3.1 Poem Classification

We evaluate a subject-based classification task on poems written by Rabindranath Tagore, who is one of the greatest poets of Bangla, into 5 categories as indicated by the author himself. The categories are পূজা ( pūjā) (devotion), প্রেম ( prēma) (love), প্রকৃতি ( prakr̥ti) (nature), স্বদেশ ( sbadēśa) (nationalism), and বিচিত্রা ( bicitrā) (miscellaneous). We collected the poems from the website of The Complete Works of Tagore available at https://tagoreweb.in/. Five-fold cross-validation was done on all 1,451 poems.

### 8.3.2 Sentiment Classification

Sentiment analysis is the task of classifying people's opinions and emotions towards entities such as products, services, organizations, and others. Sentiment classification is a prevalent downstream task in Bangla to indicate the efficacy of the corpus. For our work, we use a publicly available dataset. The dataset [20] consists of 3,307 negative and 8,500 positive reviews annotated on YouTube Bangla drama.

### 8.3.3 Spelling error detection

Spelling error is a customary problem in every language. The prevalent Bangla as "বিষ" ( biṣa, poison) versus "বিশ" ( biśa, twenty) and "কোন" ( kōna, which) versus "কোণ" ( kōṇa, angle) since both the words are present in Bangla vocabulary. The issues get greatly manifested in texts produced through OCR, such as when "সূর্য অস্ত গেল" ( sūrya asta gēla, "the sun set") gets changed to "সূর্য অন্ত গেল" ( sūrya anta gēla, "the sun ended"). In both these sentences, all the words are correct and are part of Bangla vocabulary. There is no flaw in the grammar either. To correct such errors, the model needs to understand the semantics and context of a sentence. We used the dataset created by [3] consisting of 110,356 sequence pairs. We used 80% of the data for finetuning and 20% for testing.

### 8.3.4 NLI

NLI is the task of determining whether a hypothesis is correct, wrong, or neutral with respect to a given premise sentence. It helps in understanding a model's ability to reason and handle linguistic phenomena such as negation, quantifiers, and paraphrase. We used the BNLI dataset curated by [2]. The dataset consists of 381,449 samples for training, 2,419 for validation and 4,895 samples for testing.

| Task | Vacaspati | Indic-Vac33% | Indic-Vac50% | Indic-Vac | Indic33%-Vac | Indic50%-Vac | Indic |
|---|---|---|---|---|---|---|---|
| Spell Checker | 77.72 | 75.65 | 75.22 | 76.64 | 78.18 | 77.45 | 74.58 |
| Sentiment Classification | 94.75 | 92.25 | 94.00 | 94.25 | 94.25 | 94.97 | 93.86 |
| NLI | 59.78 | 66.29 | 66.22 | 65.33 | 64.50 | 70.00 | 60.10 |
| Poem Classification | 60.39 | 50.20 | 54.35 | 60.00 | 50.68 | 49.86 | 48.46 |

**Table 15:** Performance on various downstream tasks for different model variants

Table 15 shows the performance of all the variants of the model for all the downstream tasks. From the table, it is observed that the performance of the models increases after the addition of even a fraction of VĀCASPATI on all downstream tasks compared to the model trained only on IndicCorp dataset. This result highlights the effect of quality literary data on modelling neural architectures.

# 9 Semantic Drift Analysis of Bangla Words Across Time

In this section, we delve deep into assessing how the meaning of certain words has changed over periods, specifically how their usage has evolved before and after 1941. Our aim is to investigate the socio-cultural and linguistic shifts in Bangla. Hence, we performed a semantic drift analysis focused on how the *usage and co-occurrence patterns* of words have changed across two time periods: **pre-1941** and **post-1941**. We curated a list of 150 words and grouped them under three broad categories, namely **gender** (further classified into male and female), **race/ethnicity** and **occupation**. These groupings help us understand how specific word usage has changed over time, enabling us to gauge the biases present in the corpus. 'ছেলে', 'পুরুষ', 'স্বামী', 'রাজা' ('chēlē', 'puruṣa', 'sbāmī', 'rājā') and others comprised male-gender whereas 'মেয়ে', 'মহিলা', 'স্ত্রী', 'রানী' ('mēẏē', 'mahilā', 'strī', 'rānī') comprised female-gender class. The race/ethnicity class consists of words such as 'কালো', 'বিহারী', 'মুসলমান', 'মারোয়ারী' ('kālō', 'bihārī', 'musalamāna', 'mārōẏārī') whereas 'ডাক্তার', 'নার্স', 'শিক্ষক', 'পুলিশ' ('ḍāktāra', 'nārsa', 'śikṣaka', 'puliśa') are part of occupation class. We extracted the co-occurring context words for each word within a symmetric window of size $k = 5$ in both periods. The co-occurrence distributions were then converted into normalized probability distributions. We then computed the **Jensen-Shannon Distance (JSD)** between the two context distributions for each word, producing a semantic drift score between 0 (no change) and 1 (complete divergence).

Fig 7 shows the attention map of semantic drifts of each category of words based on JSD-values. The figure shows a significant change in the usage of the words under this category, with the highest average drift for race/ethnicity class. There is significant drift in usage of words like "বাঙ্গাল", " নেপালি", " সচিব", " চাচা", " সাঁওতাল", " বিহারী", " কৃষ্ণা", " মাগি", "মুসলমান", " হিন্দু", " কালো " (``bāṅgāla'', ``nēpāli'', ``saciba'', ``cācā'', ``sām̐ōtāla'', ``bihārī'', ``kr̥ṣṇā'', ``māgi'', ``puliśa'', ``hindu'', ``kālō'') and others.

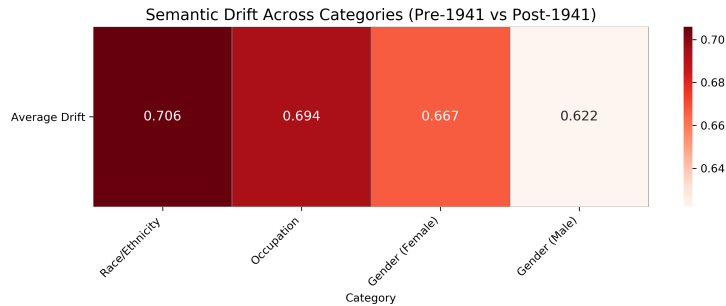

**Figure 7:** Figure showing Semantic Drift between different categories of sensitive data form pre-1941 and post-1941.

## 10 Sensitive Information

We classified email addresses, phone numbers, bank account numbers, Aadhar numbers, and PAN card numbers as sensitive information which should be protected. We followed a regular expressing matching technique where email addresses are marked with"@" followed by ".com", and Phone numbers are followed by 8-10 digits after "Ph:", "Ph.No:", or "Phone Number". Aadhar card numbers are 12-digit numbers following "Aadhar No:" or "Aadhar Number:" whereas PAN numbers are alphanumeric strings after "Pan No". The result of this study is shown in Table 16. "Actual" column of Table 16 shows the number of occurences of the string like "Aadhar" and "Phone Number", whereas valid represents the number of times the string is followed by required (8-12) digits. Newspapers contain significantly more sensitive data, posing higher security risks. Literary texts contain fewer personal identifiers, reducing exposure to privacy violations. Hence, models developed on top of them will be less susceptible to security violations. However, we have not checked the authenticity of the information used in this corpora. Our outcome is just an indication of how random scraping may impact the privacy of the models.

| Corpus | Bank Account | | Phone Number | | Aadhar | | Email | | PAN Number | |
|---|---|---|---|---|---|---|---|---|---|---|
| | Actual | Valid | Actual | Valid | Actual | Valid | Actual | Valid | Actual | Valid |
| IndicCorp | 30,392 | 2,172 | 5,130 | 78 | 3,544 | 3,499 | 0 | 0 | 4 | 0 |
| Vācaspati | 78 | 7 | 31 | 2 | 1 | 1 | 23 | 0 | 0 | 0 |

**Table 16:** Occurrence of Sensitive Information in Bangla Corpora

## 11 Impact of different Types in Vācaspati

Vācaspati contains six different types of composition, including novels, stories, essays, dramas, letters and poems. In this section, we will try to examine whether the presence of different types of composition affects the lexical diversity of Vācaspati. For this, we calculated **TTR** and **Bigram diversity** for each of the types of composition. Table 17 reports the TTR and Bigram diversity values of each type. All types of compositions have higher bigram diversity than that of the entire Vācaspati, other than novels and stories. On the other hand, all types of composition have a higher TTR value than novels. These results indicate the lexical richness of each type of composition and thus validate the use of curating a dataset with all of them. Letters are very few, which accounts for their extremely high TTR and Bigram diversity values.

We further evaluated the impact of each genre type by combining all five types and excluding one. We then calculated their TTR and Bigram diversity. Table 18 reports the values for each of the six combinations. It is observed that for all five combinations, other merging without novels, the bigram value is less than the bigram diversity value of the entire corpus of 45.13%. The TTR value of 3.53 for the entire corpus is also less for all the combinations other than novel and is similar for stories. These results underscore the impact that each type of composition has in enhancing the lexical richness of Vācaspati. The number of words and sentences is significantly higher than in other types, and novels also exhibit a more repetitive pattern compared to other types of composition. These results also highlight that curating a corpus consisting only of novels may not provide the necessary lexical diversity.

| Corpus  | TTR (%) | Bigram_diversity (%) |
|---------|---------|----------------------|
| Novels  | 3.39    | 40.29                |
| Essays  | 9.88    | 63.22                |
| Dramas  | 13.35   | 69.31                |
| Stories | 5.22    | 45.01                |
| Letters | 47.57   | 94.52                |
| Poems   | 11.69   | 68.71                |

Table 17: TTR and bigram diversity for individual text types.

| Corpus                | TTR (%) | Bigram_diversity (%) |
|-----------------------|---------|----------------------|
| merged_except_Novels  | 5.96    | 49.32                |
| merged_except_Essay   | 3.28    | 38.81                |
| merged_except_Drama   | 3.40    | 39.052               |
| merged_except_Stories | 3.63    | 40.88                |
| merged_except_Letter  | 3.40    | 39.03                |
| merged_except_Poems   | 3.33    | 38.73                |

Table 18: TTR and bigram diversity by merging five types of composition and leaving out one .

## 12  Impact of different Genres in Vācaspati

Vācaspati contains seventeen different genres of work, including children's, sci-fi, social, religion, and others. In this section, we will try to examine whether the presence of different types of genres affects the lexical diversity of Vācaspati. For this, we calculated **TTR** and **Bigram diversity** of 5 primary genres including *Children*, *Religion*, *Social*, *Scifi* and *Thriller*. We merged all other genres into one "Others" category. Table 19 reports the TTR and Bigram diversity values of each type of genre. All types of compositions have a higher TTR value than that of the entire Vācaspati, which has a TTR of 3.53. On the other hand, all 5 types of genres exhibit higher bigram diversity than the entire corpus. The "Others" class has slightly less bigram diversity compared to 45.17% of the entire dataset. These results indicate the lexical richness of each type of genre and thus validate the use of curating a dataset with all of them.

| Corpus   | TTR (%) | Bigram_diversity (%) |
|----------|---------|----------------------|
| Children | 10.65   | 65.74                |
| Religion | 12.20   | 53.35                |
| Scifi    | 9.32    | 62.88                |
| Social   | 4.00    | 43.78                |
| Thriller | 4.46    | 47.16                |
| Others   | 4.80    | 44.71                |

**Table 19:** TTR and bigram diversity for individual genres.

## 13 Conclusion

In this work, we systematically examined the lexical properties of VĀCASPATI, a literary corpus, and compared them with a much larger IndicCorp, a Bangla newspaper corpus. Across different metrics, including type–token ratio, hapax legomena ratio, and word bigram diversity—we observed that VĀCASPATI consistently exhibits greater lexical variety and morphological richness than IndicCorp, even when the corpora are compared at similar word or sentence levels. At the same time, the merged corpus analysis revealed that injecting even modest amounts of literary data into a large newspaper corpus can enrich it lexically. Global distributional analyses, including Zipf's law, entropy, redundancy, and $n$-gram perplexity, revealed the richness of information in literary corpora compared to newspaper corpora. This trend is consistent for both English and Bangla, indicating its generalizability. Our findings also reveal that perplexity of English is always less than that of Bangla, suggesting that methodologies employed for English NLP may not be suitable for Bangla. Finally, our downstream experiments validate the efficacy of VĀCASPATI, a literary corpus in NLP tasks. Lexicons induced from Vācaspati enabled more accurate detection and correction of real-word merge errors than those derived from newspaper text, and Electra models pretrained with even a fraction of Vācaspati consistently outperformed baselines trained only on IndicCorp across multiple Bangla tasks. These results highlight that high-quality literary data can substantially enhance the downstream effectiveness of Bangla language models. In future, we will try to investigate this lexical diversity of literary corpora compared to a newspaper corpora for other Indian languages and evaluate the generalizability of the observed trends.

## References

[1] Gaurav Arora. iNLTK: Natural language toolkit for indic languages. In Eunjeong L. Park, Masato Hagiwara, Dmitrijs Milajevs, Nelson F. Liu, Geeticka Chauhan, and Liling Tan, editors, *Proceedings of Second Workshop for NLP Open Source Software (NLP-OSS)*, pages 66–71, Online, Novem-

ber 2020. Association for Computational Linguistics. doi:10.18653/v1/2020.nlposs-1.10. URL https://aclanthology.org/2020.nlposs-1.10/.

[2] Abhik Bhattacharjee, Tahmid Hasan, Wasi Ahmad Uddin, Kazi Mubasshir, Md. Saiful Islam, Anindya Iqbal, M. Sohel Rahman, and Rifat Shahriyar. Banglabert: Lagnuage Model Pretraining and Benchmarks for Low-Resource Language Understanding Evaluation in Bangla. In *Findings of the North American Chapter of the Association for Computational Linguistics: NAACL 2022*, 2022. URL https://arxiv.org/abs/2101.00204.

[3] Pramit Bhattacharyya, Joydeep Mondal, Subhadip Maji, and Arnab Bhattacharya. VACASPATI: A diverse corpus of Bangla literature. In Jong C. Park, Yuki Arase, Baotian Hu, Wei Lu, Derry Wijaya, Ayu Purwarianti, and Adila Alfa Krisnadhi, editors, *Proceedings of the 13th International Joint Conference on Natural Language Processing and the 3rd Conference of the Asia-Pacific Chapter of the Association for Computational Linguistics (Volume 1: Long Papers)*, pages 1118–1130, Nusa Dua, Bali, November 2023. Association for Computational Linguistics. doi:10.18653/v1/2023.ijcnlp-main.72. URL https://aclanthology.org/2023.ijcnlp-main.72/.

[4] Meri Coleman and T. L. Liau. A computer readability formula designed for machine scoring. *Journal of Applied Psychology*, 60:283–284, 1975.

[5] Michael A. Covington and Joe D. McFall. Cutting the gordian knot: The moving-average type–token ratio (mattr). *Journal of Quantitative Linguistics*, 17(2):94–100, 2010. doi:10.1080/09296171003643098. https://doi.org/10.1080/09296171003643098. URL https://doi.org/10.1080/09296171003643098.

[6] Sumanth Doddapaneni, Rahul Aralikatte, Gowtham Ramesh, Shreya Goyal, Mitesh M. Khapra, Anoop Kunchukuttan, and Pratyush Kumar. Towards leaving no Indic language behind: Building monolingual corpora, benchmark and models for Indic languages. In *Proceedings of the 61st Annual Meeting of the Association for Computational Linguistics (Volume 1: Long Papers)*, pages 12402–12426, Toronto, Canada, July 2023. Association for Computational Linguistics. doi:10.18653/v1/2023.acl-long.693. URL https://aclanthology.org/2023.acl-long.693.

[7] Rudolf Flesch. A new readability yardstick. *Journal of Applied Psychology*, 32:221–233, 1948.

[8] Dirk Goldhahn, Thomas Eckart, and Uwe Quasthoff. Building Large Monolingual Dictionaries at the Leipzig Corpora Collection: From 100 to 200 languages. In *Proceedings of the Eighth International Conference on Language Resources and Evaluation (LREC'12)*, pages 759–765, Istanbul, Turkey, May 2012. European Language Resources Association (ELRA). URL http://www.lrec-conf.org/proceedings/lrec2012/pdf/327_Paper.pdf.

[9] Tahmid Hasan, Abhik Bhattacharjee, Kazi Samin, Masum Hasan, Madhusudan Basak, M. Sohel Rahman, and Rifat Shahriyar. Not Low-Resource Anymore: Aligner Ensembling, Batch Filtering, and New Datasets for Bengali-English Machine Translation. In *Proceedings of the 2020 Conference on Empirical Methods in Natural Language Processing (EMNLP)*, pages 2612–2623. Association for Computational Linguistics, nov 2020. doi:10.18653/v1/2020.emnlp-main.207. URL https://aclanthology.org/2020.emnlp-main.207.

[10] H. S. Heaps. *Information Retrieval: Computational and Theoretical Aspects*. Academic Press, Inc., USA, 1978. isbn:0123357500.

[11] Dan Jurafsky and James H. Martin. *Speech and Language Processing*, volume 3rd. Pearson, 2021. URL https://web.stanford.edu/~jurafsky/slp3/.

[12] Dev Kabra, Rudra Gohel, Shreyansh Prajapati, and Manish K Gupta. Statistical analysis of hindi and sanskrit languages. April 2025. doi:10.36227/techrxiv.174585786.68138641/v1. URL http://dx.doi.org/10.36227/techrxiv.174585786.68138641/v1.

[13] Divyanshu Kakwani, Anoop Kunchukuttan, Satish Golla, Gokul N.C., Avik Bhattacharyya, Mitesh M. Khapra, and Pratyush Kumar. IndicNLPSuite: Monolingual corpora, evaluation benchmarks and pre-trained multilingual language models for Indian languages. In Trevor Cohn, Yulan He, and Yang Liu, editors, *Findings of the Association for Computational Linguistics: EMNLP 2020*, pages 4948–4961, Online, November 2020. Association for Computational Linguistics. doi:10.18653/v1/2020.findings-emnlp.445. URL https://aclanthology.org/2020.findings-emnlp.445/.

[14] J. P. Kincaid, R. P. Fishburne, R. L. Rogers, and B. S. Chissom. Derivation of new readability formulas (automated readability index, fog count and flesch reading ease formula) for navy enlisted personnel. *Research Branch Report*, 1975.

[15] Anthony McEnery, Paul Baker, Rob Gaizauskas, and Hamish Cunningham. EMILLE: building a corpus of South Asian languages. In *Proceedings of the International Conference on Machine Translation and Multilingual Applications in the new Millennium: MT 2000*, University of Exeter, UK, November 20-22 2000. URL https://aclanthology.org/2000.bcs-1.11.

[16] Mohammad Mumin, Abu Awal, Md Shoeb, Mohammad Selim, and Muhammed Iqbal. Supara: A Balanced English-Bengali Parallel Corpus. 16:46–51, 01 2012.

[17] Pedro Javier Ortiz Suárez, Benoît Sagot, and Laurent Romary. Asynchronous Pipeline for Processing Huge Corpora on Medium to Low Resource Infrastructures. In Piotr Bański, Adrien Barbaresi, Hanno Biber, Evelyn Breiteneder, Simon Clematide, Marc Kupietz, Harald Lüngen, and

Caroline Iliadi, editors, *7th Workshop on the Challenges in the Management of Large Corpora (CMLC-7)*, Cardiff, United Kingdom, July 2019. Leibniz-Institut für Deutsche Sprache. doi:10.14618/IDS-PUB-9021. URL https://hal.inria.fr/hal-02148693.

[18] Ioan-Iovitz Popescu and Gabriel Altmann. Hapax legomena and language typology. *Journal of Quantitative Linguistics*, 15:370–378, 11 2008. doi: 10.1080/09296170802326699.

[19] Sagor Sarker. Bnlp: Natural language processing toolkit for Bengali language, 2021. URL https://arxiv.org/abs/2102.00405.

[20] Salim Sazzed. Cross-lingual sentiment classification in low-resource Bengali language. In *Proceedings of the Sixth Workshop on Noisy User-generated Text (W-NUT 2020)*, pages 50–60, Online, November 2020. Association for Computational Linguistics. doi:10.18653/v1/2020.wnut-1.8. URL https://aclanthology.org/2020.wnut-1.8.

[21] C. E. Shannon. A mathematical theory of communication. *The Bell System Technical Journal*, 27(3):379–423, 1948. doi:10.1002/j.1538-7305.1948.tb01338.x.

[22] Jörg Tiedemann and Lars Nygaard. The OPUS Corpus - Parallel and Free. In *Proceedings of the Fourth International Conference on Language Resources and Evaluation (LREC'04)*, Lisbon, Portugal, May 2004. European Language Resources Association (ELRA). URL http://www.lrec-conf.org/proceedings/lrec2004/pdf/320.pdf.

[23] George Kingsley Zipf. *Human Behavior and the Principle of Least Effort: An Introduction to Human Ecology*. Addison-Wesley, 1949.

# A  Appendix

## A.1  Heaps' Law

Heaps' law [10] describes the growth of vocabulary size as a function of corpus size. For the total number of tokens, $N$, in a corpus and $V(N)$ being the number of distinct word types observed, Heaps' law states that

$$V(N) = K\,N^\beta,$$

where $K > 0$ and $0 < \beta < 1$ are corpus-specific constants. $\beta$ is estimated empirically by fitting a linear regression model on the over multiple observations $(N_i, V(N_i))$ from the corpus.

## A.2  Hapax Legomena Ratio

Table A1 and Fig 8 shows the HLR of merged dataset of IndicCorp and Vācaspati. Merged dataset always have 5%-10% points more HLR than IndicCorp indicating the lexical richness of Vācaspati.

| Fraction of corpus | Literature | Newspaper | Merged |
|---|---|---|---|
| 0.1 | 4.08 | 1.02 | 1.16 |
| 0.2 | 3.22 | 0.76 | 0.87 |
| 0.3 | 2.78 | 0.63 | 0.72 |
| 0.4 | 2.51 | 0.55 | 0.62 |
| 0.5 | 2.31 | 0.50 | 0.58 |
| 0.6 | 2.15 | 0.46 | 0.52 |
| 0.7 | 2.02 | 0.42 | 0.47 |
| 0.8 | 1.91 | 0.40 | 0.43 |
| 0.9 | 1.81 | 0.38 | 0.40 |
| 1.0 | 1.72 | 0.36 | 0.39 |

**Table A1:** Variation of HLR with fraction of corpus for literary, newspaper and merged data.

## A.3  Bigram Diversity

Table A2 and Fig 9 shows the HLR of merged dataset of IndicCorp and Vācaspati. Merged dataset always have 5%-10% points more HLR than IndicCorp indicating the lexical richness of Vācaspati.

## A.4  Phonological Analysis

Phonological analysis in written corpora helps understand the latent sound structure that helps in processing and interpreting texts. Metrics like Consonant-to vowel ratio and impact the latent sound structure of a language. In Bangla, this is particularly important due to its syllabic nature and phonologically rich script. However, the modern methodologies count প্রমিত (pramita) as প্+র+ম্+ই+ত (p+ra+m+i+ta), accounting to 5 characters. Hence, we have done a Unicode correction to match with the phonological effect of words, and now প্রমিত (pramita) will result in প্+র+অ+ম্+ই+ত্+অ, accounting to 7 characters.

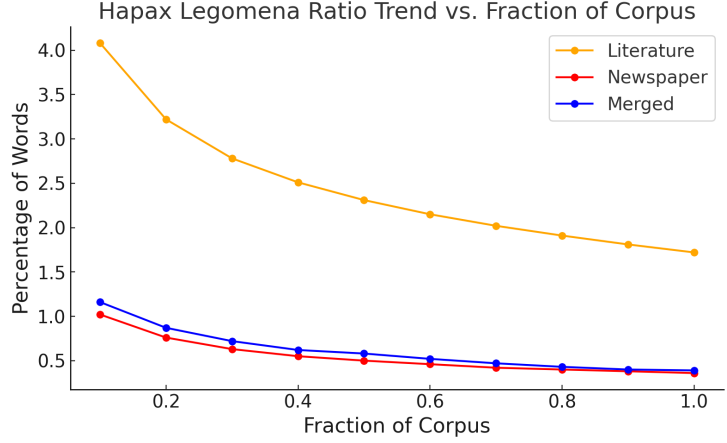

**Figure 8:** Figure showing the hapax legomena ratio of VĀCASPATI, IndicCorp and Merged corpus at different fractions of their total size.

**Consonant-to-Vowel Ratio:** The consonant-to-vowel ratio (C/V ratio) reflects the *phonotactic balance* in a corpus. Phonotactic balance in a corpus refers to the extent to which a corpus adequately represents a language's permissible sound sequences (phonotactics). [1] A lower C/V ratio indicates more vowel characters relative to consonants, contributing to smoother, more fluid articulation, resulting in the language being sweeter. In our analysis, the C/V ratio for Bangla literature is 1.542, while that of the newspaper corpus is 1.557. The result is along the expected lines since a literary corpus is more poetic than a newspaper corpus, favouring a more vowel-rich and potentially melodious structure. Further analysis revealed that newspaper corpus have more usage of words with consonant clusters like স্ম, স্ব and others that propels its C/V ratio. Words like স্মরণ (smaraṇa), স্বপ্ন (sbapna), স্বভাব (sbabhāba) have been used 102,842, 40,763, and 10,457 respectively in IndicCorp corpus. In contrast these words are only used 10,994, 7,544 and 3,379 times in the entire VĀCASPATI corpus. Our analysis shows that newspaper corpus have ∼27% higher consonant clusters than literary corpus resulting it having higher C/V ratio. However, both Bangla newspaper and literary corpus show a low C/V ratio (<2.5) which is mimicing French (Maddieson, 2005) and demonstrate Bangla being a melodious

---

[1]Native Bengali words generally have a maximum syllabic structure of CVC (consonant-vowel-consonant) and don't allow initial consonant clusters. However, for borrowed words, like স্মরণ, স্বপ্ন (smaraṇa, sbapna) initial consonant clusters are allowed.

| Fraction of corpus | Literature | Newspaper | Merged |
|---|---|---|---|
| 0.1 | 64.93 | 43.53 | 43.38 |
| 0.2 | 59.69 | 38.09 | 38.09 |
| 0.3 | 56.69 | 35.03 | 35.11 |
| 0.4 | 54.37 | 32.95 | 33.07 |
| 0.5 | 52.66 | 31.37 | 31.72 |
| 0.6 | 51.27 | 30.11 | 30.50 |
| 0.7 | 50.10 | 29.07 | 29.48 |
| 0.8 | 49.09 | 28.19 | 28.62 |
| 0.9 | 48.20 | 27.30 | 27.87 |
| 1.0 | 47.07 | 26.75 | 27.21 |

**Table A2:** Variation of bigram diversity with fraction of corpus for literary, newspaper, and merged data.

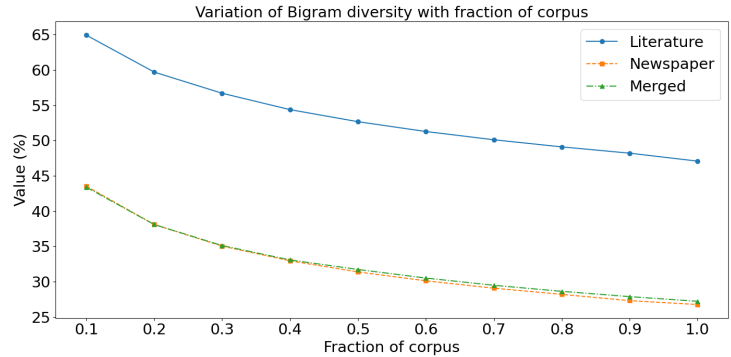

**Figure 9:** Figure showing the bigram diversity of VĀCASPATI, IndicCorp and Merged corpus at different fractions of their total size.

language.

## A.5 Zipf's Law

The top 100 words forcing the corpora to deviate from Zipf's law are provided here.

For IndicCorp, the top 100 words deviating from Zipf's law (in order of rank) are: ও, করে, না, এই, থেকে, হয়, এ, করা, এবং, তিনি, হয়েছে, হবে, তার, নিয়ে, জন্য, এক, আর, করতে, যে, একটি, এর, বলেন, হয়ে, সঙ্গে, কিন্তু, করেন, বলে, মধ্যে, কথা, যায়, দিয়ে, সেই, আমি, সময়, পর, তাদের, তা, আমার, সে, ছিল, তবে, বা, করার, ওই, হচ্ছে, তাঁর, শুরু, আমাদের, কিছু, আমরা, তারা, রয়েছে, করেছে, আছে, টাকা, পারে, নয়, পুলিশ, কাজ, সব, একটা, নেই, করেছেন, আগে, কাছে, মনে, কি, কোনো, দিন, জানান, শেষ, অনেক, এখন, বছর, বিভিন্ন, দুই, বাংলাদেশ, গত, হতে, দেখা, তাই, নতুন, প্রথম, হাজার, আজ, দেওয়া, হিসেবে, আরও, তো, দিকে, কোন, ছিলেন, বেশি, বড়, দেশের, পরে, পর্যন্ত, এমন, অভিযোগ, মতো।

ō, karē, nā, ēi, thēkē, haẏa, ē, karā, ēbaṁ, tini, haẏēchē, habē, tāra, niẏē, janya, ēka, āra, karatē, yē, ēkaṭi, ēra, balēna, haẏē, saṅgē, kintu, karēna, balē, madhyē, kathā, yāẏa, diẏē, sēi, āmi, samaẏa, para, tādēra, tā, āmāra, sē, chila, tabē, bā, karāra, ōi, hacchē, tām̐ra, śuru, āmādēra, kichu, āmarā, tārā, raẏēchē, karēchē, āchē, ṭākā, pārē, naẏa, puliśa, kāja, saba, ēkaṭā, nēi, karēchēna, āgē, kāchē, manē, ki, kōnō, dina, jānāna, śēṣa, anēka, ēkhana, bachara, bibhinna, dui, bāṁlādēśa, gata, hatē, dēkhā, tāi, natuna, prathama, hājāra, āja, dēōẏā, hisēbē, āraō, tō, dikē, kōna, chilēna, bēśi, baṛa, dēśēra, parē, paryanta, ēmana, abhiyōga, matō.

For VĀCASPATI, the top 100 words deviating from Zipf's law (in order of rank) are: না, করে, তার, আমি, আর, সে, আমার, এই, যে, একটা, থেকে, কিন্তু, ও, কি, কথা, আছে, বলল, তো, মনে, এবং, হয়ে, সঙ্গে, এক, বলে, করতে, হবে, কী, সেই, এ, নেই, কিছু, তুমি, গেল, দিকে, তা, তিনি, হয়, ছিল, আমাদের, দিয়ে, নিয়ে, তোমার, বললেন, কাছে, সব, যেন, মধ্যে, আমরা, তখন, জন্য, কেন, করা, এখন, যদি, একটু, কোনো, তাকে, আবার, পারে, এসে, হয়, হল, মতো, আমাকে, তাই, তাদের, বা, যায়, আপনি, কোন, তারপর, ঠিক, খুব, গেছে, চলে, একটি, তবে, তারা, নয়, পর, অনেক, কেউ, এর, থাকে, হাত, এমন, আপনার, হয়ে, দেখে, ওর, যা, যখন, করল, দেখা, বসে, হচ্ছে, হয়েছে, নিজের, কে, যাবে.

nā, karē, tāra, āmi, āra, sē, āmāra, ēi, yē, ēkaṭā, thēkē, kintu, ō, ki, kathā, āchē, balala, tō, manē, ēbaṁ, haẏē, saṅgē, ēka, balē, karatē, habē, kī, sēi, ē, nēi, kichu, tumi, gēla, dikē, tā, tini, haẏa, chila, āmādēra, diẏē, niẏē, tōmāra, balalēna, kāchē, saba, yēna, madhyē, āmarā, takhana, janya, kēna, karā, ēkhana, yadi, ēkaṭu, kōnō, tākē, ābāra, pārē, ēsē, haẏa, hala, matō, āmākē, tāi, tādēra, bā, yāẏa, āpani, kōna, tārapara, ṭhika, khuba, gēchē, calē, ēkaṭi, tabē, tārā, naẏa, para, anēka, kēu, ēra, thākē, hāta, ēmana, āpanāra, haẏē, dēkhē, ōra, yā, yakhana, karala, dēkhā, basē, hacchē, haẏēchē, nijēra, kē, yābē.

For Indic-Vācaspati, the top 100 words deviating from Zipf's law (in order) are: ও, করে, থেকে, এই, না, এ, হয়, করা, এবং, তার, তিনি, আর, হবে, হয়েছে, নিয়ে, এক, যে, জন্য, করতে, হয়ে, সঙ্গে, কিন্তু, একটি, এর, আমি, বলে, বললেন, কথা, সে, আমার, করেন, সেই, মধ্যে, দিয়ে, যায়, তা, পর, সময়, ছিল, তাদের, একটা, আছে, তবে, বা, কিছু, কি, আমাদের, আমরা, মনে, করার, হচ্ছে, তাঁর, নেই, ওই, তারা, সব, পারে, কাছে, নয়, করেছে, তো, রয়েছে, কোনো, কাজ, টাকা, আগে, এখন, অনেক, দিকে, করেছেন, পুলিশ, দিন, শেষ, তাই, দেখা, কী, তাকে, যদি, বছর, দুই, কোন, জানান, বিভিন্ন, আজ, মতো, নতুন, বাংলাদেশ, গত, গেছে, প্রথম, এমন, তাকে, যদি, বড়, বেশি, আরও, পরে, থাকে, হাজার, পর্যন্ত.

ō, karē, thēkē, ēi, nā, ē, haẏa, karā, ēbaṁ, tāra, tini, āra, habē, haẏēchē, niẏē, ēka, yē, janya, karatē, haẏē, saṅgē, kintu, ēkaṭi, ēra, āmi, balē, balēna, kathā, sē, āmāra, karēna, sēi, madhyē, diẏē, yāẏa, tā, para, samaẏa, chila, tādēra, ēkaṭā, āchē, tabē, bā, kichu, ki, āmādēra, āmarā, manē, karāra, hacchē, tām̐ra, nēi, ōi, tārā, saba, pārē, kāchē, naẏa, karēchē, tō, raẏēchē, kōnō, kāja, ṭākā, āgē, ēkhana, anēka, dikē, karēchēna, puliśa, dina, śēṣa, tāi, dēkhā, kī, tākē, yadi, bachara, dui, kōna, jānāna, bibhinna, āja, matō, natuna, bāṁlādēśa, gata, gēchē, prathama, ēmana, tākē, yadi, baṛa, bēśi, āraō, parē, thākē, hājāra, paryanta.

## A.6 Entropy and Redundancy of Vācaspati

Figure 10 shows the comparison of entropy and redundancy values of Vācaspati, IndicCorp and Merged corpora. Figures11 and Figure 12 shows the comparison of redundancy of Vācaspati and IndicCorp as a function of number of sentences. Vācaspati has significantly lower redundancy indicating the corpus is more diverse and information-rich.

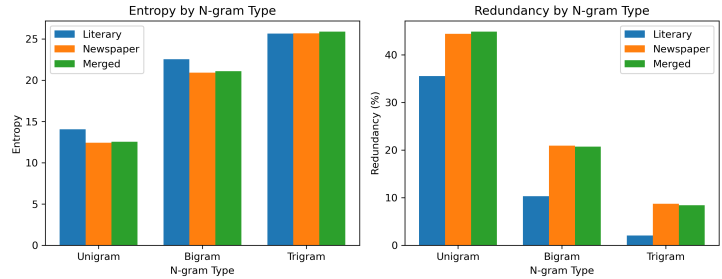

**Figure 10:** Figure showing comparison of entropy and redundancy of Vācaspati, IndicCorp and Merged datasets.

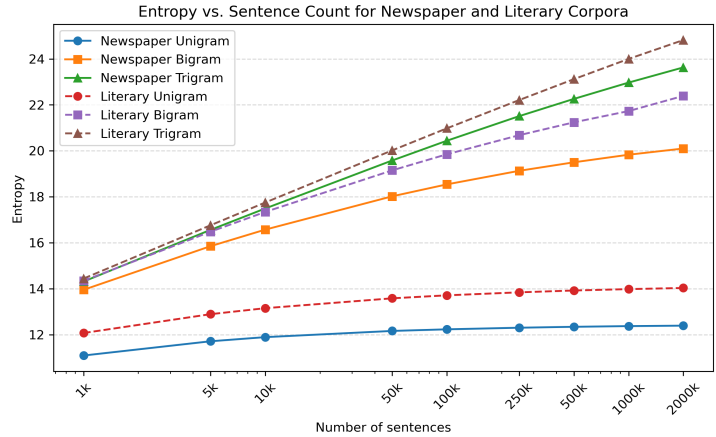

**Figure 11:** Figure showing comparison of entropy of Vācaspati and IndicCorp as a function of number of sentences.

## A.7 Perplexity

Figure 13 comparison of perplexity of $n$-gram models for Bangla and English. Perplexity of Bangla datasets are higher than that of English datasets.

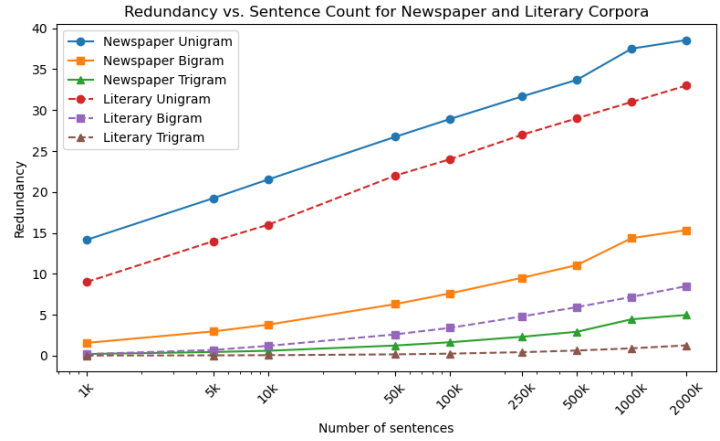

**Figure 12:** Figure showing comparison of redundancy of Vācaspati and IndicCorp as a function of number of sentences.

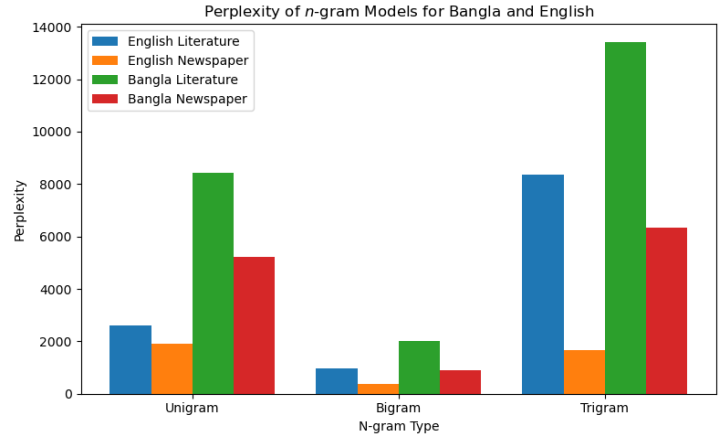

**Figure 13:** Figure showing comparison of perplexity of $n$-gram models for Bangla and English